%% file: main.tex
\documentclass{article}

\usepackage{microtype}
\usepackage{graphicx}
\usepackage{booktabs} 

\usepackage[accepted]{icml2025}

\usepackage{amsmath}
\usepackage{amssymb}
\usepackage{mathtools}
\usepackage{amsthm}

\usepackage{multirow}
\usepackage{subcaption}
\usepackage{url}

\usepackage{breakurl}
\usepackage[breaklinks]{hyperref}

\usepackage[capitalize,noabbrev]{cleveref}

\theoremstyle{plain}

\theoremstyle{definition}

\theoremstyle{remark}

\usepackage[textsize=tiny]{todonotes}

\icmltitlerunning{Adversarial Robustness Unhardening via Backdoor Attacks in Federated Learning}

\begin{document}

\twocolumn[
\icmltitle{Adversarial Robustness Unhardening via Backdoor Attacks in Federated Learning}



\icmlsetsymbol{equal}{*}

\begin{icmlauthorlist}
\icmlauthor{Taejin Kim}{equal,yyy}
\icmlauthor{Jiarui Li}{xxx}
\icmlauthor{Nikhil Madaan}{zzz}
\icmlauthor{Shubhranshu Singh}{zzz}
\icmlauthor{Carlee Joe-Wong}{zzz}

\end{icmlauthorlist}

\icmlaffiliation{yyy}{CACI International, Florham Park, New Jersey USA}
\icmlaffiliation{xxx}{University of Michigan, Michigan, USA}
\icmlaffiliation{zzz}{Carnegie Mellon University, Pittsburgh, Pennsylvania, USA}

\icmlcorrespondingauthor{Taejin Kim}{Taejin.Kim@caci.com}

\icmlkeywords{Machine Learning, ICML}

\vskip 0.3in
]




\input{sections/s0_abstract}
\input{sections/s1_introduction}
\input{sections/s2_related_works}

\input{sections/s3_FAT}

\input{sections/s4_replacement}
\input{sections/s5_defense}
\input{sections/s6_system}
\input{sections/s7_conclusion}

\newpage

\input{sections/broader}
\bibliography{references}
\bibliographystyle{icml2025}


\input{sections/s8_supplementary}

\end{document}

%% file: sections/s0_abstract.tex
\begin{abstract}
The delicate equilibrium between user privacy and the ability to unleash the potential of distributed data is an important concern. Federated learning, which enables the training of collaborative models without sharing of data, has emerged as a privacy-centric solution. This approach brings forth security challenges, notably poisoning and backdoor attacks where malicious entities inject corrupted data into the training process, as well as evasion attacks that aim to induce misclassifications at test time. Our research investigates the intersection of adversarial training, a common defense method against evasion attacks, and backdoor attacks within federated learning. We introduce Adversarial Robustness Unhardening (ARU), which is employed by a subset of adversarial clients to intentionally undermine model robustness during federated training, rendering models susceptible to a broader range of evasion attacks. We present extensive experiments evaluating ARU's impact on adversarial training and existing robust aggregation defenses against poisoning and backdoor attacks. Our results show that ARU can substantially undermine adversarial training's ability to harden models against test-time evasion attacks, and that adversaries employing ARU can even evade robust aggregation defenses that often neutralize poisoning or backdoor attacks. 
\end{abstract}

%% file: sections/s1_introduction.tex
\section{Introduction} \label{sec:introduction}

Federated learning has emerged as a promising distributed training paradigm \citep{FLmobile, fedlearn_survey} to address privacy concerns linked with the escalating growth and utilization of sensitive data generated and collected by modern computing devices, such as smartphones and Internet of Things (IoT) sensors. It allows multiple users to collaboratively train a machine learning model without the necessity of sharing their individual data. Instead, during the training phase, participants autonomously update the model, and these updates are aggregated at a central server to produce a global model sent back to the clients.
The proliferation of federated learning, and machine learning in general, has facilitated advancements in numerous applications. However, it has also ushered in a wave of attacks on learning algorithms. Evasion attacks~\citep{biggio2013evasion, madryAttacks}, for instance, manipulate inputs to trained models in ways imperceptible to human users but capable of altering the model's output during testing. For example, slight alterations to a stop sign might result in its misclassification as a speed limit sign~\citep{cao2017mitigating}. Backdoor and poisoning attacks have adversarial clients participating in model training, sending manipulated weight information during aggregation to negatively impact the trained global model \citep{xie2019dba}.

\begin{figure}[t!]
\begin{center}
\centerline{\includegraphics[width=0.99\columnwidth]{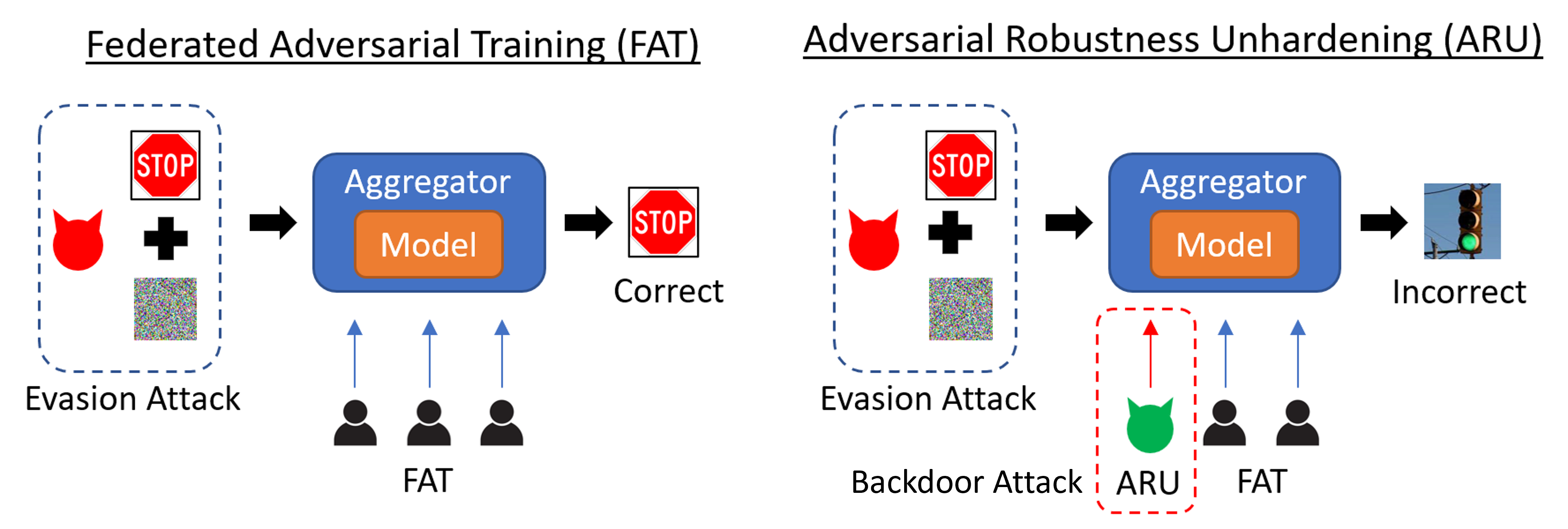}}
    \vskip -0.1in
\caption{With federated adversarial training, the clients jointly train a model robust against evasion attacks (left). However, when a small subset of train-time clients perform the ARU backdoor attack, the jointly trained model performs poorly against evasion attacks (right).}
\label{fig:ARU_demo}
\end{center}
\vskip -0.35in
\end{figure} 

\textbf{Threat Model.} Existing defenses against evasion attacks in federated learning generally utilize adversarial training~\citep{FAT, biasvar, certifiedfed, kim2022pfeddef}, where clients generate adversarial inputs and incorporate them into the training process, which has been shown to be an effective and reliable defense method against evasion attacks \citep{madryAttacks}.
However, such defenses only consider attacks generated and deployed during the testing phase of the federated learning model. In this paper, we present a \emph{novel train-time backdoor adversary that complements such test-time evasion attack adversaries by discreetly interfering with the federated adversarial training process}. As seen in Figure \ref{fig:ARU_demo}, the interfering adversary tailors its training updates to reduce the classification accuracy against evasion attacks (i.e., robustness) of the trained model, while leaving performance against benign, unaltered inputs high to avoid detection by other clients.

\textbf{Challenges and Contributions.}  
The primary challenge in implementing the proposed ARU attack lies in the fact that the attacking clients constitute only a small subset of the total client pool. Consequently, any attack strategy must operate within the constraints of the limited data and computational resources available to these attacking clients. Federated learning clients are also often resource-constrained, e.g., mobile or IoT devices with limited compute and communication resources, exacerbating the challenge of executing the ARU attack with limited client resources. 
Moreover, na\"ive ARU strategies may be susceptible to robust aggregation defenses, which have been previously developed as a defense against model poisoning attacks in federated learning~\citep{trimmedmean}. Thus, additional challenges include ensuring that ARU can bypass defense strategies that would likely be utilized in practice. 
Resource constraints, however, may also undermine clients' ability to deploy adversarial training, making the model easier to attack; thus, we aim to evaluate ARU under such realistic resource constraints.
In this paper, our main contributions are the following:

\begin{itemize}

    \item We are the first, to the best of our knowledge, to characterize and analyze the \emph{coordination between train-time (backdoor) and test-time adversaries for enhancing the success rate of evasion attacks} via Adversarial Robustness Unhardening (ARU).
    Although existing attacks (e.g., the pixel-pattern backdoor \citep{bagdasaryan2020backdoor}) have a similar concept, they instead embed a backdoor for a narrow subset of inputs, while ARU enhances the performance of \textit{all} evasion attacks.
    

    \item We propose a novel and realistically deployable method for ARU clients to obtain a non-robust model to perform a backdoor attack with. The ``extraction method'' undoes the effects of adversarial training on the global model via overfitting on curated local data, both of which are readily available to adversarial clients.
    
    \item We conduct a thorough evaluation of the ARU attack, as well as its performance against existing defense mechanisms, namely robust aggregation schemes for federated learning. We discuss the trade-offs of employing different defense methods and show that ARU attacks are still effective, even when faced with most practical defenses. We briefly examine a new defense method, that reduces privacy concerns while improving effectiveness against ARU.
\end{itemize}

The remainder of this paper is organized as follows. The Related Works section contrasts ARU with previous works. 
The Federated Adversarial Training (FAT) section introduces the FAT procedure, and evaluates its impact on test-time evasion attacks. 
The Threat Model section introduces the train-time ARU attack, its impact on the robustness of trained models, and realistic deployment methods given adversarial constraints.
In the Robust Aggregation Scheme section, we test ARU against existing robust aggregation defenses, while in the Extended Evaluation section we test ARU under varying system settings and a new defense against ARU.
We end with our Conclusion section.

This is an extended version of the work that has been presented at the NeurIPS 2023 Workshop on Backdoors in Deep Learning - The Good, the Bad, and the Ugly. 

%% file: sections/s2_related_works.tex
\section{Related Works} \label{sec:related}

\textbf{Backdoor and Poisoning Attacks.} Within the training phase of federated learning systems, malicious clients have the potential to engage in backdoor or poisoning attacks, where colluding clients send manipulated weight information during the aggregation phase with the intent of degrading the performance of the global model, either generally or for specific sub-tasks \cite{xia_poisonfl_survey}. Such attacks are generally performed by manipulating the uploaded model weights directly, or by manipulating the training data of adversarial clients.
Pixel-pattern backdoors require the attacker to modify the pixels of the digital image in a special way at test time in order for the model to misclassify the modified image due to a backdoor planted during the training phase \cite{bhagoji2019analyzing, gu2019badnets}. Although pixel-pattern backdoors are conceptually similar to the ARU attack proposed in this paper, the pixel-pattern backdoor is strictly weaker than evasion attacks as it lacks the flexibility to perform attacks with different target labels on the fly and is less agnostic to the model attacked (e.g., a partially implemented backdoor will reduce pixel-pattern attack success) \cite{bagdasaryan2020backdoor}.
To further enhance the persistence and efficacy of these attacks, \citet{xie2019dba} introduce a distributed attack strategy that employs localized triggers to poison individual attacker models while collectively exploiting the shared model. Additionally, \citet{wang2020attack} present edge-case backdoor attacks that target prediction sub-tasks unlikely to be encountered in the training or test data sets, yet still represent plausible real-world scenarios. While backdoor attacks have shown some susceptibility to defenses like adversarial training~\citep{gao2023effectiveness}, to the best of our knowledge no existing attack has shown an ability to undermine adversarial training and its ability to defend against evasion attacks.

\textbf{Defense Methods.} Robust aggregation schemes have been proposed such as the trimmed-mean defense or Krum as defense mechanisms against backdoor and poisoning attacks \cite{trimmedmean, Krum}.  Although these schemes are highly effective in countering simple or non-intelligent attacks by discarding outlier updates, they become vulnerable to exploitation by a Byzantine attacker who has knowledge of the specific aggregation scheme employed \cite{baruch2019little}. Less simplistic defense mechanisms aim to identify malicious clients by clustering and analyzing client updates \cite{foolsgold, nguyen2022flame, ozdayi2021defending}. However, more sophisticated methods often violate the privacy of individual clients and undermine the purpose of federated learning \cite{Pillutla_2022}; there exists an innate trade-off between privacy, robustness, and implementation overhead. For example, a coordinate-wise median aggregation scheme successfully defends against any model replacement attack, including ARU, but violates the privacy of clients and converges poorly given non-i.i.d. data across clients \cite{bagdasaryan2020backdoor}. Recent work \citep{ramezani2022mixtailor} has proposed to navigate these trade-offs by using a randomly chosen defense method in each iteration; this also prevents attackers from knowing which defense they will face, improving robustness~\cite{baruch2019little}. While we propose a hybrid defense against ARU that similarly combines multiple defenses, we split our defense methods across different model parameters, not across different training rounds.

%% file: sections/s3_FAT.tex
\section{Federated Adversarial Training} \label{sec:FAT}

ARU attempts to undermine federated adversarial training (FAT), which aims to train a federated learning model that is robust against evasion attacks. Adversarial training includes evasion attack data points within its training set to gain familiarity and robustness against future test-time attacks. We further introduce our experimental setup and present initial evaluations of FAT's effectiveness.

\subsection{Crafting Evasion Attacks}

A test-time adversary performs an evasion attack by altering the input $x \rightarrow x'$ to alter the model prediction for $x'$.  
In this paper, we make the assumption that both adversarial training and evasion attacks are conducted using the well-known projected gradient descent (PGD) method \cite{madryAttacks}. The PGD method is one of the most popular and effective forms of evasion attacks \cite{REN2020346}.
Here, the adversary aiming to induce any incorrect classification label iteratively updates the current input $x^t$ as:
\begin{align}\label{PGD}
    x^{t+1} = \Pi_{x+S} \left(x^t + \alpha \text{sgn}(\nabla_x L(h_{\theta}, x^t,y))\right)
\end{align}
The input $x$ with correct label $y$ is perturbed along the gradient of the model loss function $L$ with model parameters $h_{\theta}$. The step size $\alpha$ is chosen to not be too small so that an effective perturbation can be quickly found, while not too large such that effective perturbations are not omitted. The perturbation to input $x$ is then projected $(\Pi_{x + S})$ to be within the perturbation budget $S$. The perturbation budget exists such that perturbations are not obvious to detection (e.g., a heavily perturbed image may be noticed by the human eye). This budget is most often a $l_2$ or $l_\infty$ norm-ball.

In this paper, we examine white-box attacks that are characterized by the attacker having full access to the internal architecture and parameters of the victim model, which is reasonable in a federated learning setting where many participating clients own a copy of the global model. This knowledge allows the attacker to craft evasion attacks with better precision, making them generally more effective. In contrast, black-box attacks operate under the assumption that the attacker has limited or no access to the victim model's internal structure, relying on input-output interactions or a substitute model to generate adversarial examples. 
Furthermore, all evasion attacks examined are \emph{untargeted} attacks that aim to have inputs classified as any incorrect output, rather than a specific desired label in a \emph{targeted} attack. Prior research indicates that patterns of adversarial robustness in federated learning systems tend to be correlated amongst targeted and untargeted attacks \cite{kim2022pfeddef}.

\subsection{The Adversarial Training Process}

Defending against evasion attacks can be represented as a saddle point problem. In adversarial training, the primary objective is to train a model that minimizes the empirical risk associated with a classification task, even in the face of the adversary's introduction of input perturbations (e.g., through PGD as demonstrated in Equation \ref{PGD}) that maximize the loss at each data point \cite{madryAttacks}.
The objective function of adversarial training is as follows:
\begin{align}\label{eq:pFAT}
    \min_{\theta} \mathbb{E}_{(x,y)\sim \Tilde{D}} \left[\max_{\delta \in S} L_{\Tilde{D}}(h_{\theta}, x+\delta,y)\right]
\end{align}
The perturbation $\delta$ added to the data $\Tilde{D}$ is bounded within a budget $S$.
In words, we desire to find model parameters $h_\theta$ that minimize the expected maximum loss when the input $x$ is perturbed by $\delta$.
During FAT, each client incorporates adversarial examples into its  training data set using gradients from its individual model 
\cite{FAT}. These gradients are subsequently aggregated centrally following  standard federated averaging (FedAvg) \cite{fedlearn_survey}.

\input{tables/UNL_tbl1}

The effects of FAT are examined and compared to FedAvg in Table \ref{tbl:UNL_orig}. 
FAT enhances the robustness of models (i.e., classification rate against evasion attacks, denoted as Adv. Acc.)
compared to FedAvg. The data sets used and experimental setups are described in the appendix. 
Each client performs measurements on the global model with their local test data. Standard deviation values across client measurements are in parentheses for tables and represented as shaded areas in graphs. Experiments on CIFAR-10 are extensively performed in the main body of this paper, while the other data sets are analyzed in the appendix.





%% file: tables/UNL_tbl1.tex

\begin{table}[t]
\caption{Test accuracy and robustness (classification rate of evasion attacks, Adv. Acc.) for different federated learning and ARU algorithms. Standard deviation in parentheses.}
\label{tbl:UNL_orig}
\centering
\begin{tabular}{l|ll}
\hline
CIFAR 10  & Test Acc.    & Adv. Acc.    \\ \hline
FedAvg    & 0.853 (0.06) & 0.006 (0.01) \\
FAT       & 0.805 (0.06) & \textbf{0.427 (0.11)} \\
ARU       & 0.850 (0.06) & 0.004 (0.01) \\ \hline
CIFAR 100 & Test Acc.    & Adv. Acc.    \\ \hline
FedAvg    & 0.389 (0.07) & 0.007 (0.01) \\
FAT       & 0.415 (0.07) & \textbf{0.106 (0.04)} \\
ARU       & 0.389 (0.07) & 0.004 (0.01) \\ \hline
Celeba    & Test Acc.    & Adv. Acc.    \\ \hline
FedAvg    & 0.801 (0.08) & 0.000 (0.00) \\
FAT       & 0.809 (0.08) & \textbf{0.516 (0.10)} \\
ARU       & 0.822 (0.07) & 0.001 (0.01) \\ \hline
FakeNewsNet & Test Acc.    & Adv. Acc.    \\ \hline
FedAvg    & 0.782 (0.05) & 0.050 (0.02) \\
FAT       & 0.728 (0.05) & \textbf{0.380 (0.06)} \\
ARU       & 0.786 (0.06) & 0.056 (0.02) \\ \hline
\end{tabular}

\end{table}

%% file: sections/s4_replacement.tex
\section{Threat Model: Adversarial Robustness Unhardening} \label{sec:aru}

The train-time ARU adversaries aim to undermine FAT such that the trained model becomes much less robust against evasion attacks. 
\textbf{In essence, the adversaries are strategically embedding a significant backdoor into the model, effectively compromising its resilience against all forms of gradient-based adversarial evasion attacks}. The objective for such adversarial clients is:
\begin{align}
    \max_{\theta}&\; \mathbb{E}_{(x,y)\sim \Tilde{D}} \left[ L(h_{\theta},x+f(h_{\theta},x,y),y)\right] \label{eq:aru_obj}\\
    {\rm where}&\; f(h_{\theta},x,y) = \operatorname*{argmax}_{\delta \in [S]} L(h_{\theta},x+\delta,y) \label{eq:pert}\\
    {\rm s.t.}&\; \mathbb{E}_{(x,y)\sim \Tilde{D}} \left[ L(h_{\theta},x,y)\right] \leq H \label{eq:const}
\end{align}
Here, the ARU clients jointly aim to alter the global weight parameters $\theta$ such that the loss of the global model is maximized against inputs with adversarial perturbations ($f(h_{\theta},x,y)$, Eq. \ref{eq:pert}). The ARU clients further desire that test loss against benign inputs remains below a certain threshold ($H$) for the stealthy implant of the backdoor (Eq. \ref{eq:const}).

Since ARU clients are typically a minority of all federated clients, they cannot simply solve for a model satisfying Equations \ref{eq:aru_obj}--\ref{eq:const} and use this non-robust model as their local models. Instead, they must take steps to ensure that their updates to the aggregation server that are based on this non-robust model ``dominate'' updates from the other, benign clients.
Below, we first describe the model replacement method used by the ARU adversaries and discuss the subsequent adjustments made to enhance its practicality.

\subsection{The Model Replacement Method}\label{subsec:replacement-aru}

A model replacement attack is performed by replacing a legitimate global model $G$ with a backdoor model ($R$) by directly manipulating model parameters.
In the ARU case, the attack is done by one or more clients. As seen in Equation \ref{eq:replace1}, a single client $j$ performs the attack by uploading updates $U_j^{t+1}$ during aggregation, while $1/\gamma_i$ indicates the weighted contribution of the client $i$ during aggregation: 
\begin{align}\label{eq:replace1}
    R = G^t +  \sum_{i = 1}^{m} \frac{1}{\gamma_i} (U_i^{t+1} - G^t)
\end{align}
To accomplish this, attacker $j$ strategically ensures that its uploaded model  $U_j^{t+1}$ survives the aggregation across all clients by boosting the difference between desired model $R$ and global model $G^t$ based on $\gamma_j$, as seen in Equation \ref{eq:replace2}:
\begin{align}\label{eq:replace2}
    U_j^{t+1} &= \gamma_j R - (\gamma_j - 1) G^t - \sum_{i=1}^{m-1} \frac{1}{\gamma_i} (U_i^{t+1} - G^t)\\
    &\approx \gamma_j (R - G^t) + G^t \label{eq:replace3}
\end{align}
When $G_t$ is near convergence, the updates from benign participants are assumed to be close to zero ($U_i^{t+1} \approx G^t$), and are omitted in Equation \ref{eq:replace3}. \citet{bagdasaryan2020backdoor} discuss how adversary $j$ can estimate $\gamma_j$ during the training procedure, if not known ahead of time. Furthermore,  model replacement can be done jointly by multiple ARU clients. For example, if contribution $\gamma_j$ is equal amongst all adversaries, each adversary can alter its boosting rate to $\frac{\gamma}{N}$, where $N$ is the number of ARU clients. 


In Table \ref{tbl:UNL_orig}, a single adversary performs ARU by model replacement using a non-robust model trained by FedAvg. The ARU attack significantly decreases the robustness of FAT while maintaining a high test accuracy against benign inputs.
\textbf{However, it is unreasonable to assume that the adversary owns the non-robust model used for replacement ahead of time}, since the global model trained before the model replacement attack is executed will have been hardened by FAT. Thus, we next examine how a small number of collaborating clients can \emph{extract} a non-robust model from the robust global model during the training process.

\begin{figure}[t!]
\begin{center}
\centerline{\includegraphics[width=0.99\columnwidth]{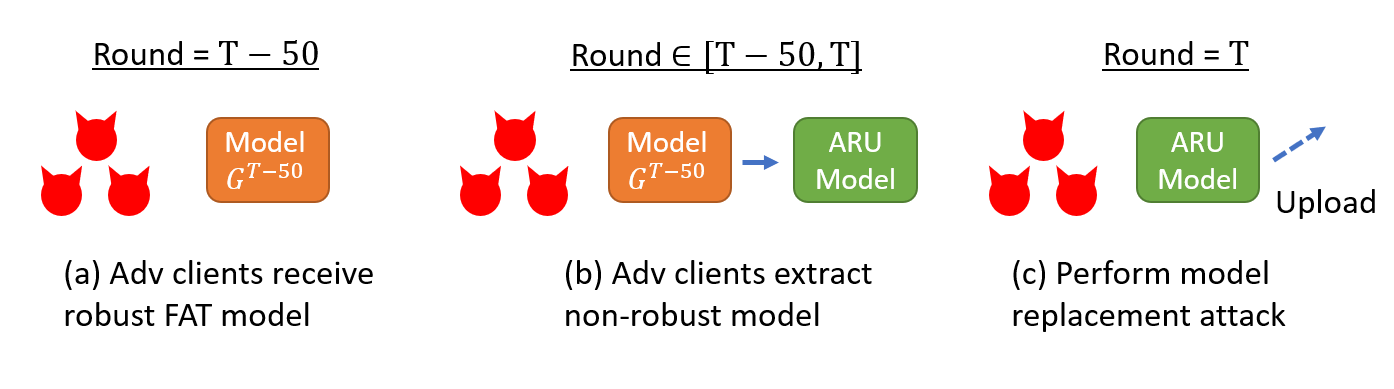}}
\caption{Extraction of a non-robust model from the global FAT model. Adversarial clients jointly weaken the FAT model and then perform a model replacement ARU attack.}
\label{fig:overfit_demo}
\end{center}
\end{figure} 

\begin{figure}[t]
    \centering
    \includegraphics[width=0.7\columnwidth]{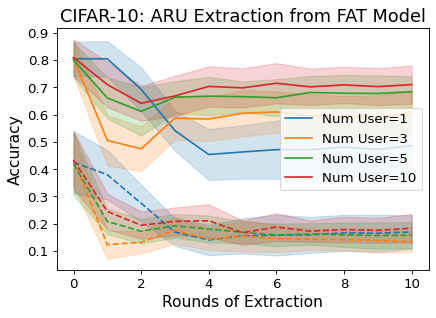}
    \caption{Extracted model performance over rounds. Test accuracy grows with more backdoor clients (solid lines), but Adv. Acc. (dashed lines) is consistently low.}
    \label{fig:overfit_result}
\end{figure}

\subsection{Extracting the Non-Robust Model for ARU}

While clients attempting to perform ARU do not reasonably have access to a non-robust model ahead of time, they do have access to the global model throughout training. We introduce \emph{a method for a small group of colluding clients to extract the non-robust model from the robust global model to then perform ARU}
In Figure \ref{fig:overfit_demo}, (a) multiple ARU clients initially receive a copy of the robust model during FAT. For a set number of rounds, (b) the ARU clients jointly update the FAT model on perturbed data. Here, data points are perturbed using PGD, and an incorrect label is assigned based on how the model classifies the perturbed data. Such manipulation and training on the ARU clients' data induces catastrophic forgetting \cite{kaushik2021understanding}, where the robustness of the global model is forgotten via overfitting on the manipulated data. Finally, (c) the adversaries jointly perform the model replacement ARU attack with the extracted model. 
In prior scenarios as presented by \citet{bagdasaryan2020backdoor}, the backdoor objective is often orthogonal to the global objective, and both can be optimized by the adversarial clients simultaneously to train the model used for replacement. However, ARU clients face a novel challenge as the global and backdoor objectives directly oppose each other and cannot be optimized simultaneously. The extraction method overcomes the conflict in objective by effectively leveraging the local data and global robust model available to obtain the model for replacement.

The performance of ARU with the extracted model is shown in Figure \ref{fig:overfit_result}. Here, different numbers of attacking clients out of 40 jointly perform the extraction for 10 rounds. The extraction is initially performed with perturbed data (50\% of data is perturbed), but only unperturbed data is used after 5 rounds to reduce the impact of the perturbed data on the test accuracy.
The lower adversarial robustness accuracy indicates the success of the extraction method, as well as the high test accuracy. 
Figure \ref{fig:overfit_result} also shows that the catastrophic forgetting of robustness occurs with few training rounds amongst the relatively small group (e.g., 1 to 5 adversaries).
With a limited number of extracting clients, the extraction procedure degrades test accuracy as the extracted model overfits to a small sample of data. 

%% file: sections/s5_defense.tex
\section{Robust Aggregation Schemes Against ARU and Backdoor Attacks} \label{sec:robust}

Robust aggregation schemes, such as trimmed-mean and median methods, are the primary defenses against poisoning and backdoor attacks \cite{yin2018byzantine}. In general, such defenses aim to discard updates from clients that stray too far from the average update, as adversarial updates tend to show high deviation from benign updates. 
In this section, we first examine the trade-off between client privacy, robustness, and overhead for different robust aggregation schemes, which informs our choice of defenses against which we evaluate ARU. Afterwards, we then observe the effect of robust aggregation schemes against the ARU attack. We first consider a single-shot attack, as described in the Threat Model section above. Informed by our experimental findings, we then propose an iterative variation of ARU and demonstrate that it can defeat common robust aggregation defenses.

\begin{figure*}[t]
    \centering
    \begin{subfigure}[t]{0.30\textwidth}
    \includegraphics[width=0.95\textwidth]{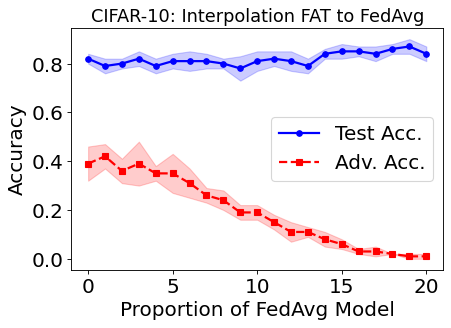}
    \caption{Robustness gradually and continuously decreases from the FAT to the FedAvg model when interpolated. Test Acc. in solid lines, Adv. Acc. in dashed lines.}
    \label{fig:fat2fedavg_gradual}
    \end{subfigure}
    \hspace{0.03cm}
    \begin{subfigure}[t]{0.33\textwidth}
    \includegraphics[width=0.97\textwidth]{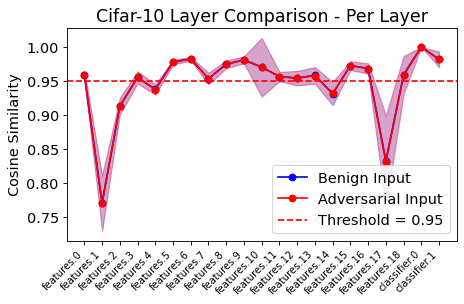}
    \caption{The cosine similarity between outputs of each layer between a FedAvg and FAT model. Each layer for both models has the previous output of a FedAvg layer as inputs.}
    \label{fig:cosine_independent}
    \end{subfigure}
    \hspace{0.03cm}
    \begin{subfigure}[t]{0.33\textwidth}
    \includegraphics[width=0.97\textwidth]{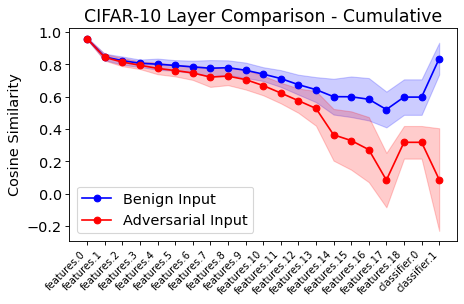}
    \caption{The cosine similarity between outputs of each layer between a FedAvg model and FAT model are compared. Differences in layer outputs are accumulated through layers.}
    \label{fig:cosine_cumulative}
    \end{subfigure}
  \caption{ The FAT and FedAvg model are highly similar. Thus, a gradual insertion of the ARU backdoor across multiple rounds is possible as robustness is continuous with respect to the FAT and FedAvg models. (CIFAR-10)}
  \label{fig:group_plot_1}
\end{figure*}

\subsection{Trade-offs of Robust Aggregation} \label{subsec:tradeoff-robust}

There exists an inherent trade-off between robustness, privacy, and overhead of defensive aggregation schemes, and it is only possible to only pursue two of the three desired characteristics during federated learning \cite{Pillutla_2022}. Any aggregation scheme that updates the global model as a linear function of client updates (e.g., FedAvg, trimmed-mean) is both privacy preserving and communication efficient, but non-robust. Alternative robust aggregation schemes \cite{foolsgold, nguyen2022flame, ozdayi2021defending} violate privacy constraints as individual updates from clients must be examined and compared. However, by significantly increasing the per-round communication cost between clients, it is possible to obfuscate individual client updates by mixing updates between a subset of clients.

Since the median method has poor privacy guarantees, which are a primary motivator of federated learning in the first place, most of our results in this section focus on evaluating ARU against the more private \textit{trimmed-mean defense}. Recent work, however, has proposed a state-of-the-art \textit{mixture} method \citep{ramezani2022mixtailor} that rotates among different defense methods in different federated learning iterations, thus allowing for a more flexible trade-off between robustness, privacy, and overhead. 
We moreover show that in Section \ref{sec:noniid_robust}, in highly non-i.i.d. settings, robust aggregation methods perform poorly even without the ARU attack, due to their inability to effectively accommodate heterogeneous client updates.

\subsection{Single-Shot ARU Against Robust Aggregation}\label{subsec:single-robust}

\input{tables/robust_tbl}

The single-shot ARU attack, where one or more clients attempt to replace the robust global model with a non-robust model as in Equation \ref{eq:replace1}, is performed against the trimmed-mean and median robust-aggregation schemes in Table \ref{tbl:robust_singleshot}. The trimmed-mean method discards $tm \in [0,1]$ proportion of largest and smallest values for each weight parameter (e.g., $tm=0.05$ discards 10\% of updates), while the coordinate-wise median method selects the median value for each individual parameter.

The comparison of Table \ref{tbl:UNL_orig} and Table \ref{tbl:robust_singleshot} indicate that the trimmed-mean method ($tm = 0.05$) mitigates the effect of the single-shot ARU attack significantly across all data sets, maintaining similar to slightly lower robustness measurements to FAT models. The median defense displays high resilience towards the single-shot ARU attack as well. 
However, as discussed in the prior subsection and by \citet{Pillutla_2022}, the coordinate-wise median method violates the privacy constraints of federated learning. Furthermore, ARU clients are not restricted to one-shot attacks, and the motivation and implementation for a multi-round iterative ARU attack is discussed in the next subsection.

\begin{figure*}[t]
    \centering
    \begin{subfigure}[t]{0.32\textwidth}
    \includegraphics[width=0.97\textwidth]{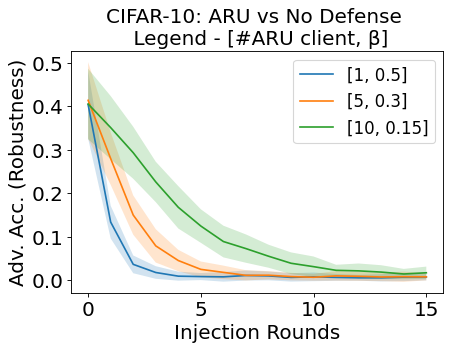}
    \caption{Performance of iterative ARU against no defense.}
    \label{fig:ARU_nodef}
    \end{subfigure}
    \hspace{0.03cm}
    \begin{subfigure}[t]{0.32\textwidth}
    \includegraphics[width=0.98\textwidth]{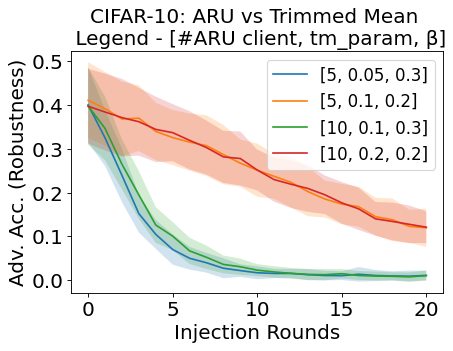}
    \caption{Performance of iterative ARU against trimmed-mean defense.}
    \label{fig:ARU_trimmedmean}
    \end{subfigure}
    \hspace{0.03cm}
    \begin{subfigure}[t]{0.32\textwidth}
    \includegraphics[width=0.98\textwidth]{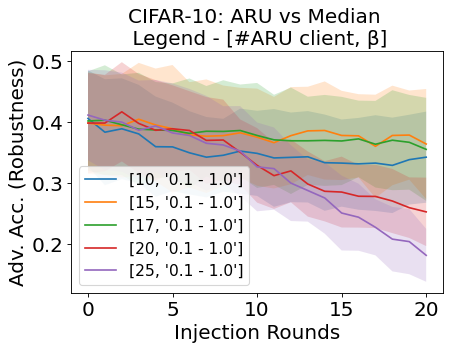}
    \caption{Performance of iterative ARU against median defense.}
    \label{fig:ARU_median}
    \end{subfigure}
    \hspace{0.03cm}
  \caption{Performance of iterative ARU against various robust aggregation schemes. Iterative ARU handily bypasses the trimmed-mean defense that fulfills both privacy and communication overhead constraints. In all cases, test accuracy for benign input remains high (above 0.79). (CIFAR-10)}
  \label{fig:group_plot_2}
\end{figure*}

\subsection{Performance of Iterative ARU Against Robust Aggregation Schemes}\label{subsec:iterative-robust}

The model replacement method used by the single-shot ARU demonstrated in Equations \ref{eq:replace1} -- \ref{eq:replace3}, calculates the backdoor updates assuming that the global model is near convergence and updates from non-ARU clients are near zero. However, when ARU clients perform an iterative attack across rounds, the assumption of near convergence is no longer valid as benign clients simultaneously aim to increase the robustness of the global model. For the iterative ARU attack to succeed, the adversarial clients must successfully estimate and counteract the uploads of the benign clients.

In Figure \ref{fig:fat2fedavg_gradual}, we analyze the robustness of interpolations between a robust FAT model and a non-robust FedAvg model at different points. The results suggest that iterative ARU can progressively erode robustness over multiple rounds as the FAT model gradually shifts toward the FedAvg model.
Figure \ref{fig:cosine_independent} measures the cosine similarity of layer outputs between FAT and FedAvg models for the same input. Most layers exhibit high similarity (above 0.95), suggesting that the two models produce nearly identical representations. This resemblance may further incentivize an iterative ARU attack, as attackers benefit when the backdoored model closely mirrors the global model.
Robustness appears to be concentrated in layers with low similarity, specifically features 1 and 17 (both convolutional layers). In Figure \ref{fig:cosine_cumulative}, we measure cosine similarity between layer outputs with only the first layer receiving identical input. The results indicate that small differences in early layers propagate, leading to significant discrepancies for adversarial inputs.


Iterative ARU is tested against no robust aggregation scheme in Figure \ref{fig:ARU_nodef}.  The attack greatly diminishes robustness of the global model. A new scaling variable $\beta$ is introduced to flexibly scale the iterative model replacement process (i.e., $U_j^{t+1}\approx \gamma_j \beta (R - G^t) + G^t$, based on Eq. \ref{eq:replace2}). As $\beta$ is decreased, the replacement of the robust global model becomes more gradual. Iterative ARU is tested against the trimmed-mean defense in Figure \ref{fig:ARU_trimmedmean}. 
ARU quickly overcomes the defense given either a lower trimmed-mean cutoff parameter $tm = 0.05$, or a higher number of ARU clients (10 clients compared to 5). Against a more aggressive cutoff policy (e.g., $tm = 0.2$), lowering the parameter $\beta$ aids in bypassing the defense (for CIFAR-100 in Figure \ref{fig:group_plot_9}), as adversarial updates are closer to the expected global updates.
In all circumstances, iterative ARU is implemented against a model thus far trained via FAT that has had no robust aggregation performed up until the point where the ARU attack is started.



Iterative ARU is tested against the median aggregation scheme in Figure \ref{fig:ARU_median}. Here, each client begins with $\beta = 0.1$ and increases the parameter by $0.1$ until it reaches $1$. In general, the median defense is robust and can only be bypassed if the number of adversarial clients is over half of the total number of federated clients (i.e., greater than 20 of 40 clients). However, possibly due to the proximity of the ARU update to the average global update induced by the constraint of Equation \ref{eq:const}, ARU demonstrates a slight reduction in robustness even with less than half of clients as adversaries. Moreover, as discussed above it may raise privacy concerns.

%% file: tables/robust_tbl.tex
\begin{table}[t]
\caption{Robust aggregation defense against one-shot ARU attacks. Attacks are performed by a single client for a single round. Trimmed-mean cutoff set at $tm=0.05$.}
\label{tbl:robust_singleshot}
\centering
\begin{tabular}{l|l|ll}
\hline
Dataset                    & Metric    & Trim-Mean    & Median       \\ \hline
\multirow{2}{*}{CIFAR10}   & Test Acc. & 0.811 (0.07) & 0.809 (0.07) \\
                           & Adv. Acc. & 0.423 (0.09) & 0.416 (0.09) \\ \hline
\multirow{2}{*}{CIFAR100}  & Test Acc. & 0.417 (0.06) & 0.419 (0.06) \\
                           & Adv. Acc. & 0.099 (0.04) & 0.093 (0.03) \\ \hline
\multirow{2}{*}{Celeba}    & Test Acc. & 0.796 (0.08) & 0.786 (0.08) \\
                           & Adv. Acc. & 0.465 (0.11) & 0.481 (0.10) \\ \hline
\multirow{2}{*}{FakeNews}  & Test Acc. & 0.697 (0.07) & 0.670 (0.07) \\
                           & Adv. Acc. & 0.369 (0.06) & 0.386 (0.05) \\ \hline
\end{tabular}
\end{table}

%% file: sections/s6_system.tex
\begin{figure*}[t]
    \centering
    \begin{subfigure}[t]{0.31\textwidth}
    \includegraphics[width=0.95\textwidth]{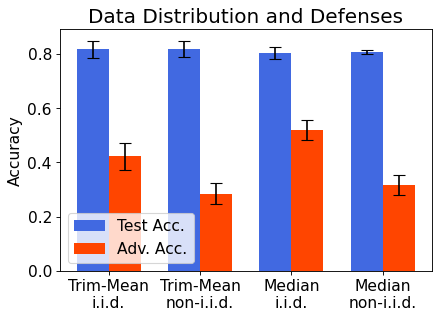}
    \caption{FAT trained model performance given different data distributions (due to resource constraints) and aggregation schemes.}
    \label{fig:noniid_tm}
    \end{subfigure}
    \hspace{0.03cm}
    \begin{subfigure}[t]{0.34\textwidth}
    \includegraphics[width=0.99\textwidth]{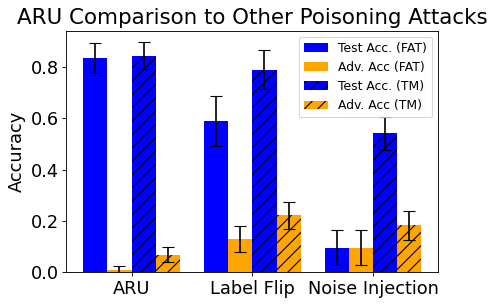}
    \caption{Comparison of iterative ARU against alternative multi-round poisoning attacks for FAT, with and without trimmed-mean defense. Trimmed mean cutoff set at $tm = 0.1$.}
    \label{fig:vs_other_poison}
    \end{subfigure}
    \hspace{0.03cm}
    \begin{subfigure}[t]{0.31\textwidth}
    \includegraphics[width=0.90\textwidth]{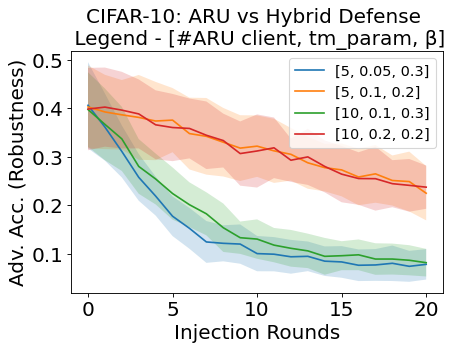}
    \caption{Hybrid defense scheme against iterative ARU that combines median with trimmed-mean.}
    \label{fig:hybrid_defense}
    \end{subfigure}
    \hspace{0.03cm}
  \caption{Extended experiments on CIFAR-10. (a) ARU is more effective in non-i.i.d. data settings. ARU (b) is more effective than other attacks at inducing mis-classifications and (c) has its effectiveness not fully but significantly reduced by the proposed hybrid defense of trimmed-mean and median methods.}
  \label{fig:group_plot_3}
\end{figure*}

\section{Extended Evaluation: ARU in Realistic Systems} \label{sec:system}

We evaluate ARU beyond the results of previous sections. 
We first examine the effects of limited system resources and of different data distributions across clients. Afterwards, we compare ARU to other poisoning attacks and demonstrate a potential defense against ARU that reduces privacy violations compared to the median defense. 

\subsection{Distributed System Resources and  Non-i.i.d. Data} \label{sec:noniid_robust}

As discussed by \citet{bagdasaryan2020backdoor} and \citet{Pillutla_2022}, non-i.i.d. data distributions amongst federated clients reduce the efficacy of robust aggregation schemes by making even benign model updates more heterogeneous. The non-i.i.d. data challenge is further exacerbated by the implementation of FAT amongst resource constrained clients, as the uneven generation of adversarial examples amongst the federated client training set further differentiates data, and thus model update, characteristics \cite{kim2022pfeddef,bnprop}.

The effects of non-i.i.d. data against robust aggregation schemes are examined in Figure \ref{fig:noniid_tm}. Data is split in a more i.i.d. manner by setting the proportion of adversarial examples equivalent across all clients (40\% of each client training set). The non-i.i.d. split is induced by limiting the number of clients (15 of 40) that can generate adversarial training examples, while on average maintaining the total number of adversarial training points in the system. While all of our experiments use non-i.i.d. data distributions (see the appendix for exact settings), this result shows that some distributions, especially in relation to available resources for adversarial training, can cause even the median defense to lose effectiveness.
For both trimmed-mean and median methods, the overall robustness of a FAT model is 
significantly reduced in the non-i.i.d. compared to i.i.d. case (14\% and 20\% reductions, respectively). Thus, although robust aggregation schemes display some level of robustness against the ARU attack, they may still reduce the robustness of a model trained by FAT in realistic settings.

\subsection{Varying Attack Types and Parameters}

The performance of ARU is compared against the label flip and random noise poisoning attacks \cite{pmlr-v20-biggio11}. Adversarial participants of federated adversarial training perform the label flip attack by training over data with incorrect labels, while the random noise attack is performed by generating random updates for aggregation. Both poisoning methods achieve the ARU objective of Equation \ref{eq:pFAT}, as diminishing the overall quality of the global model reduces classification accuracy against adversarial examples as well. When compared to ARU in Figure \ref{fig:vs_other_poison}, the poisoning attacks display relatively poorer performance in diminishing robustness against a trimmed-mean defense. As the constraint of Equation \ref{eq:const} is not fulfilled by the poisoning attacks, their updates tend to deviate further from the global average than the ARU attack, leading to a lower success rate in penetrating the robust aggregation scheme.

\subsection{Hybrid Defense Against ARU}

\input{tables/hybrid_defense}

We demonstrate a defense mechanism that combines elements of the trimmed-mean and median defense. As seen in Figure \ref{fig:cosine_independent}, robust FAT models and non-robust FedAvg models are only highly different in a few layers. 

We propose a hybrid defense method, that performs the median aggregation for such highly influential layers (layers 1, 2, 4, 14, 17), and trimmed-mean for the rest. In Figure \ref{fig:hybrid_defense} and Table \ref{tbl:hybrid_adv}, it is shown that performing median aggregation on just 5 of 21 layers 
significantly reduces the effectiveness against equivalent iterative ARU attacks performed against trimmed-mean defenses from Figure \ref{fig:ARU_trimmedmean}. The privacy violation of the regular median aggregation is significantly reduced, as a smaller subset of the layers has their updates from clients examined by the aggregator. 

%% file: tables/hybrid_defense.tex
\begin{table}[t]
\caption{Adv. Acc. measurement comparing trimmed-mean and the hybrid defense against equivalent iterative ARU adversaries after 20 rounds of attack. Test accuracy against benign inputs remain stable between 0.78 - 0.83 across all measurements. (CIFAR-10)}
\label{tbl:hybrid_adv}
\centering
\begin{tabular}{l|l|l|l|l}
\hline
\begin{tabular}[c]{@{}l@{}}ARU\\ Clients\end{tabular} & $tm$ & $\beta$ & \begin{tabular}[c]{@{}l@{}}Trim-Mean\\ Adv. Acc\end{tabular} & \begin{tabular}[c]{@{}l@{}}Hybrid\\ Adv. Acc.\end{tabular} \\ \hline
5                                                     & 0.05 & 0.3     & 0.011 (0.01)                                          & \textbf{0.078 (0.03)}                                               \\
5                                                     & 0.1  & 0.2     & 0.120 (0.04)                                          & \textbf{0.225 (0.06)}                                               \\
10                                                    & 0.1  & 0.3     & 0.010 (0.01)                                          & \textbf{0.081 (0.03)}                                               \\
10                                                    & 0.2  & 0.2     & 0.121 (0.04)                                          & \textbf{0.237 (0.04)}                                               \\ \hline
\end{tabular}
\end{table}

%% file: sections/s7_conclusion.tex
\section{Conclusion} \label{sec:conc}

Federated learning has a vulnerability to a multitude of adversaries due to the participation of numerous clients in both training and testing phases. In response to test-time evasion attacks, federated adversarial training has emerged as a promising defense mechanism. Our contribution, Adversarial Robustness Unhardening (ARU), represents a novel train-time counterpart to test-time evasion attacks, discreetly undermining the robustness of federated adversarial training. We demonstrate ARU's effectiveness in reducing model robustness, even when benign clients attempt to deploy robust aggregation mechanisms to reduce the impact of malicious model updates. We further show that in realistic settings with resource constraints and highly non-i.i.d. data, ARU can be even more effective as defenses like the median aggregation, which can effectively defend against ARU in other settings, achieve poor robustness in these more realistic settings.
Our work contributes to ongoing efforts to secure federated learning against adversarial threats while advancing the understanding of its intricate dynamics.

%% file: sections/s8_supplementary.tex
\newpage
\appendix
\onecolumn

\section{Appendix}

\input{tables/variable_tbl}

\subsection{Resources and Assets} \label{app:ext_rsrc}

The code that performs a demo of ARU has been attached to this submission. 
The federated learning implementation is based on the code from \cite{fedMTL21} found at \url{https://github.com/omarfoq/FedEM}. Our work has adjusted the work presented in FedEM with the following changes:
\begin{itemize}
    \item We introduce the adversarial training mechanisms for different types of distributed learning.
    \item Implement the robust aggregation schemes of trimmed-mean, median, Krum and the hybrid defense of trimmed-mean and median. 
    \item We add implementations of the ARU attack, the label flip, and noise injection attack.
\end{itemize}

\input{tables/exp_setting_tbl}

\subsection{Data Sets and Experimental Settings}\label{app:data_exp}

All experiments in the main body and appendix are carried out on an AWS EC2 instance of type g4dn.xlarge. These instance types have NVIDIA GPUs as well as CUDA.

\textbf{CIFAR.} The CIFAR-10 and CIFAR-100 data sets \cite{ref:cifar} are selected to analyze the trends of ARU for two similar classification tasks of varying sizes. Both the CIFAR-10 and CIFAR-100 models are trained on \texttt{MobileNetV2}.
As seen in Table \ref{tbl:exp_setting}, the training parameters in both instances for performing standard federated learning (FedAvg) and Federated Adversarial Training (FAT) are equivalent, except that the number in the clients in the system is 50 for CIFAR-100 compared to 40 for CIFAR-10. When performing FAT for both data sets, a fraction of the training across clients is altered based on the projected-gradient-descent (PGD) method. The perturbation budget is higher when performing adversarial training ($S = 4.5$) as compared to when the trained model is subjected to adversarial attacks ($S = 4$). Adversaries performing evasion attacks usually have a smaller perturbation budget, as their attacks must be more discrete to avoid detection. The data is artificially split between clients in a non-i.i.d. manner for both data sets with $\zeta = 0.4$. For just the CIFAR datasets, when performing iterative ARU, the number of rounds used to train the robust model is set to 150 instead, as in realistic systems the iterative ARU attack would begin before full convergence.

\textbf{Celeba.} The Celeba data set is a large-scale data set with celebrity images, each with 40 binary labels, from LEAF, a benchmarking framework for federated learning \cite{leaf}. This data set is selected for analysis because the distribution of data across clients follows a more realistic pattern than artificial division of data amongst clients used for other data sets. We combine 4 binary classification tasks (Smiling, Male, Eyeglasses, Wearing Hat) to formulate a classification problem with 16 classes. The images are reshaped to $45\times55\times3$ shaped tensors to reduce GPU and memory load. The Celeba model is trained on the \texttt{MobileNetV2} as well, with similar training and attack parameters to CIFAR-10, as seen in Table \ref{tbl:exp_setting}. 

\textbf{FakeNewsNet.} The FakeNewsNet data set consists of news articles classified as falsehood or truth by Politifact and GossipCop \cite{shu2019fakenewsnetdatarepositorynews}. For this binary classification task, each text is converted to an embedding using the BERT-based \texttt{all-mpnet-base-v2} model \cite{reimers2019sentencebertsentenceembeddingsusing}. From there, a shallow neural network is trained to classify each embedding as true or false, with the model consisting of three convolution layers, followed by two linear layers. Data points are randomly assigned to one of 20 clients. As seen in Table \ref{tbl:exp_setting}, the perturbation amount for adversarial examples are significantly reduced compared to other data sets, as the attribute size of the text embeddings are smaller than images. In practice, the perturbations in text can be generated by contrastive optimization, where specific phrases are injected into the text to significantly alter embeddings in desired manners \cite{xue2024badragidentifyingvulnerabilitiesretrieval}.

\textbf{Non-i.i.d. Data Distribution.} The data distribution process across clients is taken from \cite{fedMTL21}. When dividing data across clients during experiments, the parameter $\zeta>0$ impacts how the data is distributed. The data division process begins with an assumption $M$ underlying distributions. The underlying distributions are constructed by having each label in the data set divided in a i.i.d. manner into one of the distributions. Afterwards, data points are mapped from each distribution to all clients using the Dirichlet distribution, which takes $\zeta$ as an input parameter. When $\zeta$ is a low value, data is more non-i.i.d. across clients as there is higher variance between clients for the number of data points assigned from a specific underlying distribution. When $\zeta$ is a higher value, clients tend to have a similar number of data points from each underlying distribution compared to other clients, making the global data distribution more i.i.d.. For all experiments in the paper, the number of underlying distributions assumed is $M=3$. The seed for the non-iid data split is set to ``$12345$'' for all data sets.

\begin{figure*}[t]
    \centering
    \begin{subfigure}[t]{0.31\textwidth}
    \includegraphics[width=0.99\textwidth]{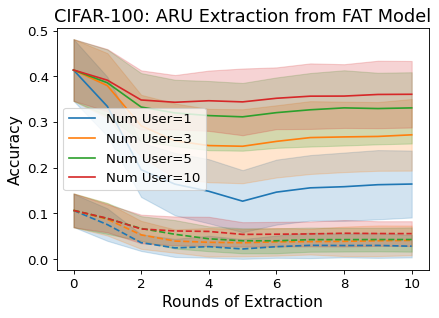}
    \end{subfigure}
    \hspace{0.03cm}
    \begin{subfigure}[t]{0.31\textwidth}
    \includegraphics[width=0.99\textwidth]{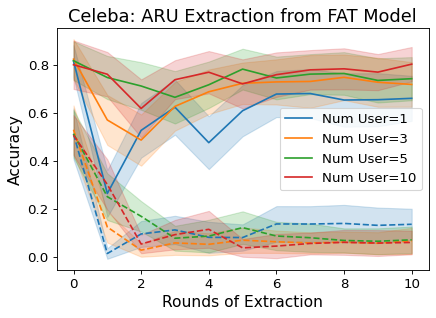}
    \end{subfigure}
    \hspace{0.03cm}
    \begin{subfigure}[t]{0.33\textwidth}
    \includegraphics[width=0.99\textwidth]{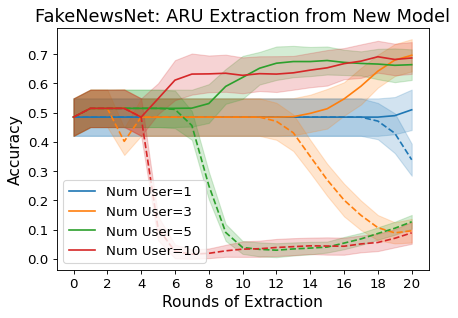}
    \end{subfigure}
    \hspace{0.03cm}
  \caption{Extracted model performance over rounds. For FakeNewsNet, rather than extracting from a FAT model, a model is trained from scratch. Generally, test accuracy grows with more backdoor clients (solid lines), but Adv. Acc. (dashed lines) is consistently low.}
  \label{fig:group_plot_4}
\end{figure*}

\begin{figure*}[t]
    \centering
    \begin{subfigure}[t]{0.32\textwidth}
    \includegraphics[width=0.99\textwidth]{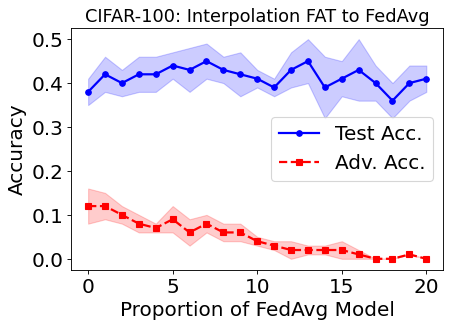}
    \end{subfigure}
    \hspace{0.03cm}
    \begin{subfigure}[t]{0.32\textwidth}
    \includegraphics[width=0.99\textwidth]{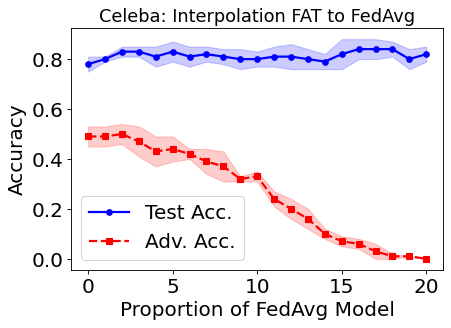}
    \end{subfigure}
    \hspace{0.03cm}
    \begin{subfigure}[t]{0.32\textwidth}
    \includegraphics[width=0.99\textwidth]{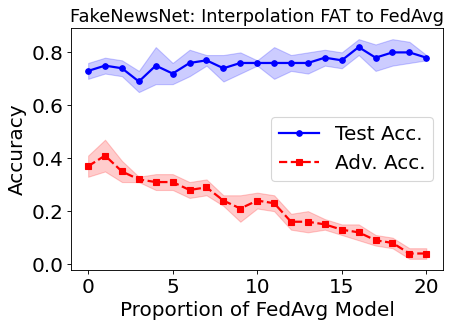}
    \end{subfigure}
    \hspace{0.03cm}
  \caption{Gradual interpolation between FedAvg and FAT models. Gradual change indicates that iterative ARU should be effective against robust global models.}
  \label{fig:group_plot_5}
\end{figure*}

\begin{figure*}[t]
    \centering
    \begin{subfigure}[t]{0.32\textwidth}
    \includegraphics[width=0.99\textwidth]{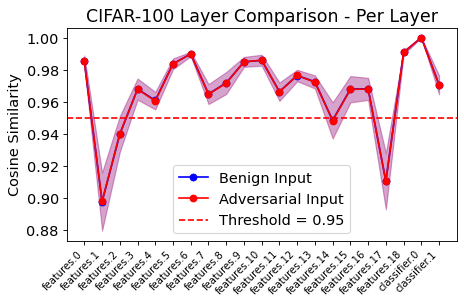}
    \end{subfigure}
    \hspace{0.03cm}
    \begin{subfigure}[t]{0.32\textwidth}
    \includegraphics[width=0.99\textwidth]{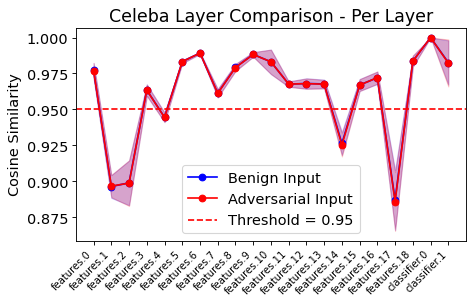}
    \end{subfigure}
    \hspace{0.03cm}
    \begin{subfigure}[t]{0.32\textwidth}
    \includegraphics[width=0.99\textwidth]{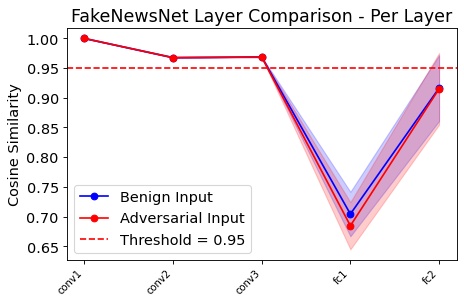}
    \end{subfigure}
    \hspace{0.03cm}
  \caption{The cosine similarity between outputs of each layer between a FedAvg model and FAT model are compared. Each layer for both models has the previous output of a FedAvg layer as inputs. Overall, a small subset of layers are highly different between FedAvg and FAT, while the majority of them are similar. Robustness is embedded in these diverging layers.}
  \label{fig:group_plot_6}
\end{figure*}

\begin{figure*}[t]
    \centering
    \begin{subfigure}[t]{0.32\textwidth}
    \includegraphics[width=0.99\textwidth]{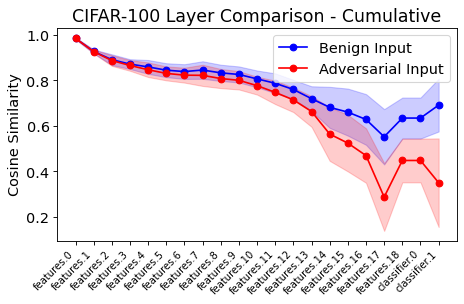}
    \end{subfigure}
    \hspace{0.03cm}
    \begin{subfigure}[t]{0.32\textwidth}
    \includegraphics[width=0.99\textwidth]{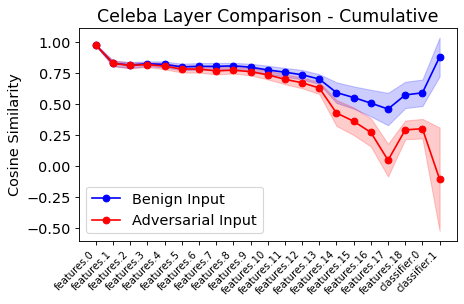}
    \end{subfigure}
    \hspace{0.03cm}
    \begin{subfigure}[t]{0.32\textwidth}
    \includegraphics[width=0.99\textwidth]{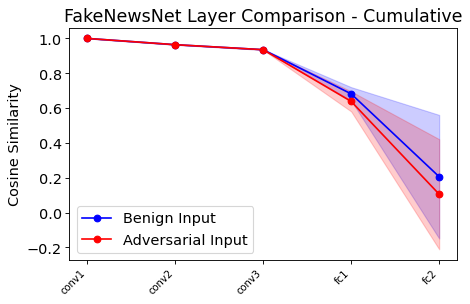}
    \end{subfigure}
    \hspace{0.03cm}
  \caption{The cosine simlarity between outputs of each layer between a FedAvg model and FAT model are compared. Differences in layer outputs are accumulated through layers, leading the similarities between benign inputs and adversarial inputs to diverge as it traverses through the layers when comparing FedAvg models to that of FAT.}
  \label{fig:group_plot_7}
\end{figure*}

\input{tables/swap_tbl}

\begin{figure*}[t]
    \centering
    \begin{subfigure}[t]{0.32\textwidth}
    \includegraphics[width=0.99\textwidth]{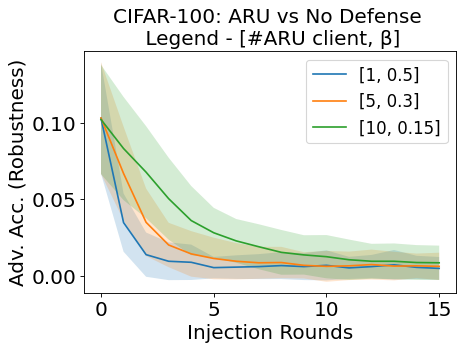}
    \end{subfigure}
    \hspace{0.03cm}
    \begin{subfigure}[t]{0.32\textwidth}
    \includegraphics[width=0.99\textwidth]{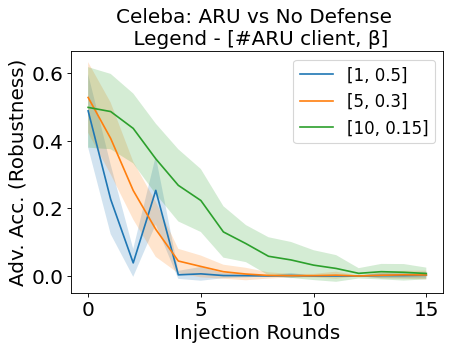}
    \end{subfigure}
    \hspace{0.03cm}
    \begin{subfigure}[t]{0.32\textwidth}
    \includegraphics[width=0.99\textwidth]{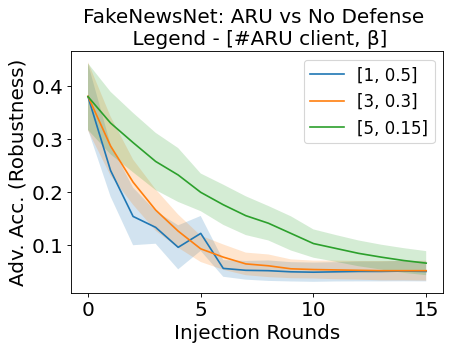}
    \end{subfigure}
    \hspace{0.03cm}
  \caption{Iterative ARU against no defense. The attack effectively reduces robustness of the global model in a few rounds.}
  \label{fig:group_plot_8}
\end{figure*}

\begin{figure*}[t]
    \centering
    \begin{subfigure}[t]{0.32\textwidth}
    \includegraphics[width=0.99\textwidth]{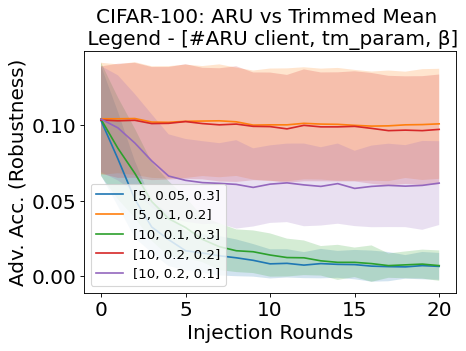}
    \end{subfigure}
    \hspace{0.03cm}
    \begin{subfigure}[t]{0.32\textwidth}
    \includegraphics[width=0.99\textwidth]{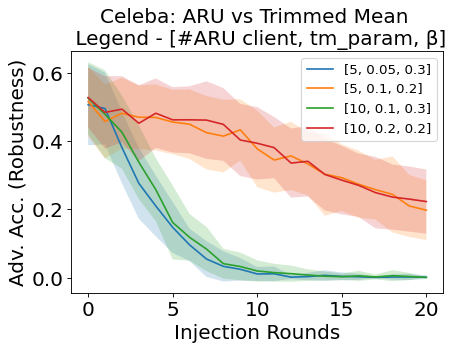}
    \end{subfigure}
    \hspace{0.03cm}
    \begin{subfigure}[t]{0.32\textwidth}
    \includegraphics[width=0.99\textwidth]{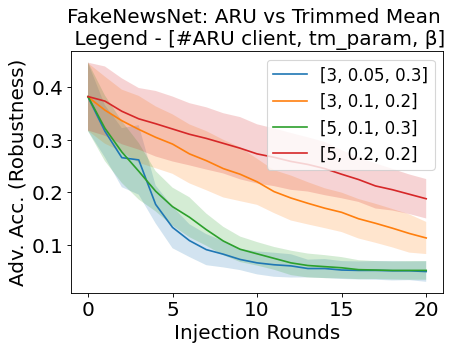}
    \end{subfigure}
    \hspace{0.03cm}
  \caption{Iterative ARU is effective against trimmed-mean defense.}
  \label{fig:group_plot_9}
\end{figure*}

\begin{figure*}[t]
    \centering
    \begin{subfigure}[t]{0.32\textwidth}
    \includegraphics[width=0.99\textwidth]{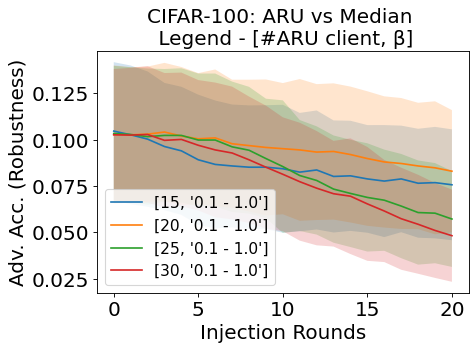}
    \end{subfigure}
    \hspace{0.03cm}
    \begin{subfigure}[t]{0.32\textwidth}
    \includegraphics[width=0.99\textwidth]{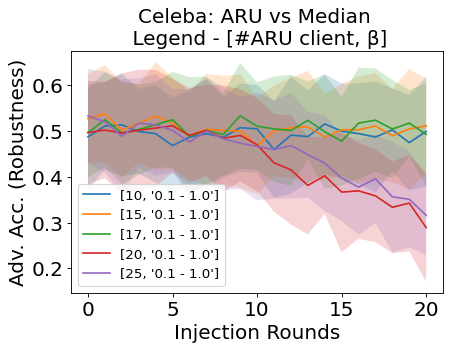}
    \end{subfigure}
    \hspace{0.03cm}
    \begin{subfigure}[t]{0.32\textwidth}
    \includegraphics[width=0.99\textwidth]{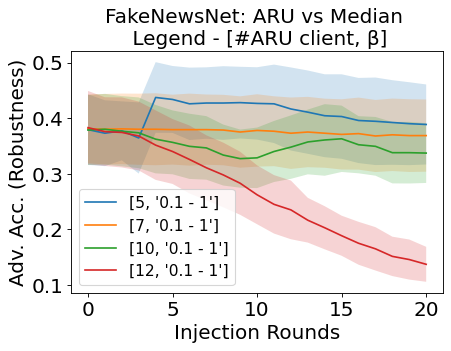}
    \end{subfigure}
    \hspace{0.03cm}
  \caption{ARU against median defense. Overall, robustness is maintained as long as the ARU clients are in the minority.}
  \label{fig:group_plot_10}
\end{figure*}


\subsection{Experimental Results for Other Data Sets}

Results for Figure \ref{fig:overfit_result}, \ref{fig:group_plot_1}, and \ref{fig:group_plot_2} are replicated for the CIFAR-100, Celeba, and FakeNewsNetwork data sets. 

\textbf{ARU Model Extraction From Robust FAT Model.}
In Figure \ref{fig:group_plot_4}, model extraction is performed by ARU clients for each data set in order to retrieve a non-robust model to perform ARU with. For CIFAR-100 and Celeba, the extraction is initially performed with perturbed data (30\% for CIFAR-100 and 50\% for Celeba), but only unperturbed data is used after 5 rounds to reduce the impact of extraction on test accuracy. The lower adversarial robustness indicates the success of the extraction method, as well as the high test accuracy. For FakeNewsNet, extraction via catastrophic forgetting is not as effective, as solutions potentially degenrate to a 50\% accuracy in a binary label scenario. Thus, a new model is trained from scratch by the adversarial clients. With few adversarial clients (e.g., 3 of 20), a non-robust model can be discovered due to the lower dimensionality of the data.

\textbf{Similarity Between FedAvg and FAT Models.}
In Figure \ref{fig:group_plot_5}, the models for each data set is gradually interpolated between the FAT and FedAvg model. In all cases, the reduction in robustness is gradual while the test accuracy against benign inputs remains high. In Figure \ref{fig:group_plot_6}, the cosine similarity is measured between outputs of layers between FAT and FedAvg models given the same input for layers for each model. It is shown that the majority of layers between FedAvg and FAT have similar outputs given the same input (above 0.95 cosine similarity). The gradual interpolation and similarity between layers motivates iterative ARU implementations. Furthermore, we can surmise that robustness is embedded with layers with low similarity. In Figure \ref{fig:group_plot_7}, the cosine simliarity is measured between outputs of layers between FAT and FedAvg models, but only the input to the first layer is the same. Here, it is observed that the difference in embedded robustness of specific layers propagate to lead to significantly different results for adversarial inputs. Swapping out these robust layers of FAT models with that of FedAvg significantly reduce the robustness as seen in Table \ref{tbl:swap} --- further solidifying the notion that robustness is embedded in a small subset of layers. 
For CIFAR-10, CIFAR-100, and Celeba 5 of 21 layers are swapped, while for FakeNewsNetwork a single layer out of 5 is swapped.

\textbf{Performance of Iterative ARU and Defenses.} 
In Figure \ref{fig:group_plot_8}, iterative ARU is tested against no robust aggregation scheme across data sets. In each of these cases, iterative ARU greatly reduces robustness of the FAT-based global model. 
Iterative ARU is then tested against the trimmed-mean defense in Figure \ref{fig:group_plot_9}. The adjustment of the scaling parameter $\beta$, that scales down the difference between the non-robust model and the robust global model to bypass defense, aids significantly especially for CIFAR-100, where reducing $\beta$ from $0.2$ to $0.1$ allows 10 adversaries to somewhat bypass the aggressive $tm = 0.2$ cutoff to reduce the robustness of the global model. 

Against the median defense, as seen in Figure \ref{fig:group_plot_10}, iterative ARU is only effective in reducing robustness when the majority of clients in the system are adversarial. In the case of CIFAR-100, even when adversarial ARU clients are in the minority, robustness is still slightly diminished, possibly due to the proximity between the FedAvg and FAT models despite their difference in robustness. 


%% file: tables/variable_tbl.tex
\begin{table}[H]
\label{tbl:vars}
\caption{Variables used to describe ARU and run relevant experiments.}
\centering
\begin{tabular}{l|l|l|l}
\hline
Variable           & Explanation                                                                                                                 & Variable   & Explanation                                                                                                                                               \\ \hline
$(x,y)\sim D$      & \begin{tabular}[c]{@{}l@{}}Data including attributes ($x$) \\ and label ($y$) drawn from \\ distribution $D$\end{tabular}   & $G^t$      & \begin{tabular}[c]{@{}l@{}}Global federated learning model \\ at round $t$ of training\end{tabular}                                                       \\ \hline
$S$                & \begin{tabular}[c]{@{}l@{}}Perturbation budget when \\ performing evasion attack\end{tabular}                               & $U_i^{t}$  & \begin{tabular}[c]{@{}l@{}}Model injected by client $i$ in \\ the system to perform model \\ replacement attack\end{tabular}                              \\ \hline
$\alpha$           & \begin{tabular}[c]{@{}l@{}}Step-size when performing \\ evasion attack\end{tabular}                                         & $\gamma_i$ & \begin{tabular}[c]{@{}l@{}}Aggregation constant of client \\ $i$ in terms of weighted averaging \\ in federated learning round\end{tabular}               \\ \hline
$h_{\theta}$       & \begin{tabular}[c]{@{}l@{}}Model parameter of globally\\ trained federated learning model\end{tabular}                      & $R$        & \begin{tabular}[c]{@{}l@{}}Desired model to be injected by \\ adversarial clients performing ARU \\ (model injection)\end{tabular}                        \\ \hline
$\delta, f(\cdot)$ & \begin{tabular}[c]{@{}l@{}}Represents the perturbation \\ added to attribute $x$ for an \\ evasion attack\end{tabular}      & $\zeta$    & \begin{tabular}[c]{@{}l@{}}Data split parameter \\ to induce non-i.i.d. setting\end{tabular}                                                              \\ \hline
$H$                & \begin{tabular}[c]{@{}l@{}}Constant representing threshold \\ which test loss can not exceed \\ for ARU attack\end{tabular} & $\beta$    & \begin{tabular}[c]{@{}l@{}}ARU attack scaling parameter, where\\ adversarial updates are scaled by \\ $\beta$ to stay closer to global model\end{tabular} \\ \hline
$tm$               & \begin{tabular}[c]{@{}l@{}}Cut-off parameter for trimmed-\\ mean aggregation\end{tabular}                                   &            &                                                                                                                                                           \\ \hline
\end{tabular}

\end{table}

%% file: tables/exp_setting_tbl.tex
\begin{table}[ht]
\caption{Settings used for federated learning for different datasets, as well as for federated adversarial training and projected-gradient-descent (PGD) attacks.}
\label{tbl:exp_setting}
\centering
\begin{tabular}{l|l|l|l}
\hline
Data Set  & Train Setting            & FAT Setting                               & Adversarial Attack Setting   \\ \hline
CIFAR-10  & 200 Rounds               & Train set $S = 4.5$, L2 norm              & Train set $S = 4$, L2 norm   \\
          & 40 Clients               & Adv. data proportion = 0.5                & PGD Steps = 10               \\
          & Learning rate = 0.01     & PGD Steps = 10, step size $\alpha$= 0.01  & Step size $\alpha$ = 0.01    \\
          & Data Split $\zeta = 0.4$ & Update Adv. data = 10 rounds              &                              \\ \hline
CIFAR-100 & 200 Rounds               & Train set $S = 4.5$, L2 norm              & Train set $S = 4$, L2 norm   \\
          & 50 Clients               & Adv. data proportion = 0.5                & PGD Steps = 10               \\
          & Learning rate = 0.01     & PGD Steps = 10, step size $\alpha$ = 0.01 & Step size $\alpha$ = 0.01    \\
          & Data Split $\zeta = 0.4$ & Update Adv. data = 10 rounds              &                              \\ \hline
Celeba    & 100 Rounds               & Train set $S = 4.5$, L2 norm              & Train set $S = 4$, L2 norm   \\
          & 40 Clients               & Adv. data proportion = 0.5                & PGD Steps = 10               \\
          & Learning rate = 0.01     & PGD Steps = 10, step size $\alpha$ = 0.01 & Step size $\alpha$ = 0.01    \\
          & Data Split $\zeta = 0.4$ & Update Adv. data = 10 rounds              &                              \\ \hline
FakeNews- & 100 Rounds               & Train set $S = 0.2$, L2 norm              & Train set $S = 0.1$, L2 norm \\
Network   & 20 Clients               & Adv. data proportion = 0.5                & PGD Steps = 10               \\
          & Learning rate = 0.01     & PGD Steps = 10, step size $\alpha$ = 0.01 & Step size $\alpha$ = 0.01    \\
          &                          & Update Adv. data = 10 rounds              &                              \\ \hline
\end{tabular}
\end{table}

%% file: tables/swap_tbl.tex

\begin{table}[ht]
\caption{Swapping out convolution layers [1, 2, 4, 14, 17] (features.1, features.3, features.4, etc.) of robust \texttt{MobileNetV2} FAT model with non-robust FedAvg model for CIFAR-10, CIFAR-100, and Celeba. A single linear layer (Fc1) is swapped out for the FakeNewsNetwork dataset instead.}
\label{tbl:swap}
\centering
\begin{tabular}{l|l|ll}
\hline
Dataset                    & Metric    & FAT          & Swapped               \\ \hline
\multirow{2}{*}{CIFAR10}   & Test Acc. & 0.805 (0.06) & 0.572 (0.07)          \\
                           & Adv. Acc. & 0.427 (0.11) & \textbf{0.083 (0.04)} \\ \hline
\multirow{2}{*}{CIFAR100}  & Test Acc. & 0.415 (0.07) & 0.279 (0.05)          \\
                           & Adv. Acc. & 0.106 (0.04) & \textbf{0.008 (0.01)} \\ \hline
\multirow{2}{*}{Celeba}    & Test Acc. & 0.809 (0.08) & 0.766 (0.06)         \\
                           & Adv. Acc. & 0.516 (0.10) & \textbf{0.101 (0.05} \\ \hline
\multirow{2}{*}{FakeNewsNetwork} & Test Acc. & 0.728 (0.05) & 0.781 (0.05)          \\
                           & Adv. Acc. & 0.380 (0.06) & \textbf{0.146 (0.03)} \\ \hline
\end{tabular}

\end{table}

%% file: main.bbl
\begin{thebibliography}{35}
\providecommand{\natexlab}[1]{#1}
\providecommand{\url}[1]{\texttt{#1}}
\expandafter\ifx\csname urlstyle\endcsname\relax
  \providecommand{\doi}[1]{doi: #1}\else
  \providecommand{\doi}{doi: \begingroup \urlstyle{rm}\Url}\fi

\bibitem[Bagdasaryan et~al.(2020)Bagdasaryan, Veit, Hua, Estrin, and Shmatikov]{bagdasaryan2020backdoor}
Bagdasaryan, E., Veit, A., Hua, Y., Estrin, D., and Shmatikov, V.
\newblock How to backdoor federated learning.
\newblock In \emph{International conference on artificial intelligence and statistics}, pp.\  2938--2948. PMLR, 2020.

\bibitem[Baruch et~al.(2019)Baruch, Baruch, and Goldberg]{baruch2019little}
Baruch, G., Baruch, M., and Goldberg, Y.
\newblock A little is enough: Circumventing defenses for distributed learning.
\newblock \emph{Advances in Neural Information Processing Systems}, 32, 2019.

\bibitem[Bhagoji et~al.(2019)Bhagoji, Chakraborty, Mittal, and Calo]{bhagoji2019analyzing}
Bhagoji, A.~N., Chakraborty, S., Mittal, P., and Calo, S.
\newblock Analyzing federated learning through an adversarial lens, 2019.

\bibitem[Biggio et~al.(2011)Biggio, Nelson, and Laskov]{pmlr-v20-biggio11}
Biggio, B., Nelson, B., and Laskov, P.
\newblock Support vector machines under adversarial label noise.
\newblock In Hsu, C.-N. and Lee, W.~S. (eds.), \emph{Proceedings of the Asian Conference on Machine Learning}, volume~20 of \emph{Proceedings of Machine Learning Research}, pp.\  97--112, South Garden Hotels and Resorts, Taoyuan, Taiwain, 14--15 Nov 2011. PMLR.
\newblock URL \url{https://proceedings.mlr.press/v20/biggio11.html}.

\bibitem[Biggio et~al.(2013)Biggio, Corona, Maiorca, Nelson, {\v{S}}rndi{\'c}, Laskov, Giacinto, and Roli]{biggio2013evasion}
Biggio, B., Corona, I., Maiorca, D., Nelson, B., {\v{S}}rndi{\'c}, N., Laskov, P., Giacinto, G., and Roli, F.
\newblock Evasion attacks against machine learning at test time.
\newblock In \emph{Joint European conference on machine learning and knowledge discovery in databases}, pp.\  387--402. Springer, 2013.

\bibitem[Blanchard et~al.(2017)Blanchard, El~Mhamdi, Guerraoui, and Stainer]{Krum}
Blanchard, P., El~Mhamdi, E.~M., Guerraoui, R., and Stainer, J.
\newblock Machine learning with adversaries: Byzantine tolerant gradient descent.
\newblock In Guyon, I., Luxburg, U.~V., Bengio, S., Wallach, H., Fergus, R., Vishwanathan, S., and Garnett, R. (eds.), \emph{Advances in Neural Information Processing Systems}, volume~30. Curran Associates, Inc., 2017.
\newblock URL \url{https://proceedings.neurips.cc/paper/2017/file/f4b9ec30ad9f68f89b29639786cb62ef-Paper.pdf}.

\bibitem[Caldas et~al.(2019)Caldas, Duddu, Wu, Li, Konečný, McMahan, Smith, and Talwalkar]{leaf}
Caldas, S., Duddu, S. M.~K., Wu, P., Li, T., Konečný, J., McMahan, H.~B., Smith, V., and Talwalkar, A.
\newblock Leaf: A benchmark for federated settings, 2019.

\bibitem[Cao \& Gong(2017)Cao and Gong]{cao2017mitigating}
Cao, X. and Gong, N.~Z.
\newblock Mitigating evasion attacks to deep neural networks via region-based classification.
\newblock In \emph{Proceedings of the 33rd Annual Computer Security Applications Conference}, pp.\  278--287, 2017.

\bibitem[Fung et~al.(2020)Fung, Yoon, and Beschastnikh]{foolsgold}
Fung, C., Yoon, C. J.~M., and Beschastnikh, I.
\newblock The limitations of federated learning in sybil settings.
\newblock In \emph{23rd International Symposium on Research in Attacks, Intrusions and Defenses (RAID 2020)}, pp.\  301--316, San Sebastian, October 2020. USENIX Association.
\newblock ISBN 978-1-939133-18-2.
\newblock URL \url{https://www.usenix.org/conference/raid2020/presentation/fung}.

\bibitem[Gao et~al.(2023)Gao, Wu, Zhang, Gan, Xia, Niu, and Sugiyama]{gao2023effectiveness}
Gao, Y., Wu, D., Zhang, J., Gan, G., Xia, S.-T., Niu, G., and Sugiyama, M.
\newblock On the effectiveness of adversarial training against backdoor attacks.
\newblock \emph{IEEE Transactions on Neural Networks and Learning Systems}, 2023.

\bibitem[Gu et~al.(2019)Gu, Dolan-Gavitt, and Garg]{gu2019badnets}
Gu, T., Dolan-Gavitt, B., and Garg, S.
\newblock Badnets: Identifying vulnerabilities in the machine learning model supply chain, 2019.

\bibitem[Hong et~al.(2021)Hong, Wang, Wang, and Zhou]{bnprop}
Hong, J., Wang, H., Wang, Z., and Zhou, J.
\newblock Federated robustness propagation: Sharing adversarial robustness in federated learning.
\newblock \emph{arXiv preprint arXiv:2106.10196}, 2021.

\bibitem[Kaushik et~al.(2021)Kaushik, Gain, Kortylewski, and Yuille]{kaushik2021understanding}
Kaushik, P., Gain, A., Kortylewski, A., and Yuille, A.
\newblock Understanding catastrophic forgetting and remembering in continual learning with optimal relevance mapping, 2021.

\bibitem[Kim et~al.(2023)Kim, Singh, Madaan, and Joe-Wong]{kim2022pfeddef}
Kim, T., Singh, S., Madaan, N., and Joe-Wong, C.
\newblock Characterizing internal evasion attacks in federated learning.
\newblock In Ruiz, F., Dy, J., and van~de Meent, J.-W. (eds.), \emph{Proceedings of The 26th International Conference on Artificial Intelligence and Statistics}, volume 206 of \emph{Proceedings of Machine Learning Research}, pp.\  907--921. PMLR, 25--27 Apr 2023.
\newblock URL \url{https://proceedings.mlr.press/v206/kim23a.html}.

\bibitem[Krizhevsky et~al.(2009)Krizhevsky, Nair, and Hinton]{ref:cifar}
Krizhevsky, A., Nair, V., and Hinton, G.
\newblock Cifar-10 (canadian institute for advanced research).
\newblock 2009.
\newblock URL \url{http://www.cs.toronto.edu/~kriz/cifar.html}.

\bibitem[Li et~al.(2021)Li, Wen, Wu, Hu, Wang, Li, Liu, and He]{fedlearn_survey}
Li, Q., Wen, Z., Wu, Z., Hu, S., Wang, N., Li, Y., Liu, X., and He, B.
\newblock A survey on federated learning systems: Vision, hype and reality for data privacy and protection.
\newblock \emph{IEEE Transactions on Knowledge and Data Engineering}, pp.\  1–1, 2021.
\newblock ISSN 2326-3865.
\newblock \doi{10.1109/tkde.2021.3124599}.
\newblock URL \url{http://dx.doi.org/10.1109/TKDE.2021.3124599}.

\bibitem[Lim et~al.(2020)Lim, Luong, Hoang, Jiao, Liang, Yang, Niyato, and Miao]{FLmobile}
Lim, W. Y.~B., Luong, N.~C., Hoang, D.~T., Jiao, Y., Liang, Y.-C., Yang, Q., Niyato, D., and Miao, C.
\newblock Federated learning in mobile edge networks: A comprehensive survey.
\newblock \emph{IEEE Communications Surveys \& Tutorials}, 22\penalty0 (3):\penalty0 2031--2063, 2020.

\bibitem[Madry et~al.(2017)Madry, Makelov, Schmidt, Tsipras, and Vladu]{madryAttacks}
Madry, A., Makelov, A., Schmidt, L., Tsipras, D., and Vladu, A.
\newblock Towards deep learning models resistant to adversarial attacks.
\newblock \emph{arXiv preprint arXiv:1706.06083}, 2017.

\bibitem[Marfoq et~al.(2021)Marfoq, Neglia, Bellet, Kameni, and Vidal]{fedMTL21}
Marfoq, O., Neglia, G., Bellet, A., Kameni, L., and Vidal, R.
\newblock Federated multi-task learning under a mixture of distributions.
\newblock \emph{Advances in Neural Information Processing Systems}, 34, 2021.

\bibitem[Nguyen et~al.(2022)Nguyen, Rieger, De~Viti, Chen, Brandenburg, Yalame, M{\"o}llering, Fereidooni, Marchal, Miettinen, et~al.]{nguyen2022flame}
Nguyen, T.~D., Rieger, P., De~Viti, R., Chen, H., Brandenburg, B.~B., Yalame, H., M{\"o}llering, H., Fereidooni, H., Marchal, S., Miettinen, M., et~al.
\newblock $\{$FLAME$\}$: Taming backdoors in federated learning.
\newblock In \emph{31st USENIX Security Symposium (USENIX Security 22)}, pp.\  1415--1432, 2022.

\bibitem[Ozdayi et~al.(2021)Ozdayi, Kantarcioglu, and Gel]{ozdayi2021defending}
Ozdayi, M.~S., Kantarcioglu, M., and Gel, Y.~R.
\newblock Defending against backdoors in federated learning with robust learning rate.
\newblock In \emph{Proceedings of the AAAI Conference on Artificial Intelligence}, volume~35, pp.\  9268--9276, 2021.

\bibitem[Pillutla et~al.(2022)Pillutla, Kakade, and Harchaoui]{Pillutla_2022}
Pillutla, K., Kakade, S.~M., and Harchaoui, Z.
\newblock Robust aggregation for federated learning.
\newblock \emph{IEEE Transactions on Signal Processing}, 70:\penalty0 1142–1154, 2022.
\newblock ISSN 1941-0476.
\newblock \doi{10.1109/tsp.2022.3153135}.
\newblock URL \url{http://dx.doi.org/10.1109/TSP.2022.3153135}.

\bibitem[Ramezani-Kebrya et~al.(2022)Ramezani-Kebrya, Tabrizian, Faghri, and Popovski]{ramezani2022mixtailor}
Ramezani-Kebrya, A., Tabrizian, I., Faghri, F., and Popovski, P.
\newblock Mixtailor: Mixed gradient aggregation for robust learning against tailored attacks.
\newblock \emph{arXiv preprint arXiv:2207.07941}, 2022.

\bibitem[Reimers \& Gurevych(2019)Reimers and Gurevych]{reimers2019sentencebertsentenceembeddingsusing}
Reimers, N. and Gurevych, I.
\newblock Sentence-bert: Sentence embeddings using siamese bert-networks, 2019.
\newblock URL \url{https://arxiv.org/abs/1908.10084}.

\bibitem[Ren et~al.(2020)Ren, Zheng, Qin, and Liu]{REN2020346}
Ren, K., Zheng, T., Qin, Z., and Liu, X.
\newblock Adversarial attacks and defenses in deep learning.
\newblock \emph{Engineering}, 6\penalty0 (3):\penalty0 346--360, 2020.
\newblock ISSN 2095-8099.
\newblock \doi{https://doi.org/10.1016/j.eng.2019.12.012}.
\newblock URL \url{https://www.sciencedirect.com/science/article/pii/S209580991930503X}.

\bibitem[Shu et~al.(2019)Shu, Mahudeswaran, Wang, Lee, and Liu]{shu2019fakenewsnetdatarepositorynews}
Shu, K., Mahudeswaran, D., Wang, S., Lee, D., and Liu, H.
\newblock Fakenewsnet: A data repository with news content, social context and spatialtemporal information for studying fake news on social media, 2019.
\newblock URL \url{https://arxiv.org/abs/1809.01286}.

\bibitem[Wang et~al.(2020)Wang, Sreenivasan, Rajput, Vishwakarma, Agarwal, Sohn, Lee, and Papailiopoulos]{wang2020attack}
Wang, H., Sreenivasan, K., Rajput, S., Vishwakarma, H., Agarwal, S., Sohn, J.-y., Lee, K., and Papailiopoulos, D.
\newblock Attack of the tails: Yes, you really can backdoor federated learning.
\newblock \emph{Advances in Neural Information Processing Systems}, 33:\penalty0 16070--16084, 2020.

\bibitem[Xia et~al.(2023)Xia, Chen, Yu, and Ma]{xia_poisonfl_survey}
Xia, G., Chen, J., Yu, C., and Ma, J.
\newblock Poisoning attacks in federated learning: A survey.
\newblock \emph{IEEE Access}, 11:\penalty0 10708--10722, 2023.
\newblock \doi{10.1109/ACCESS.2023.3238823}.

\bibitem[Xie et~al.(2019)Xie, Huang, Chen, and Li]{xie2019dba}
Xie, C., Huang, K., Chen, P.-Y., and Li, B.
\newblock Dba: Distributed backdoor attacks against federated learning.
\newblock In \emph{International conference on learning representations}, 2019.

\bibitem[Xue et~al.(2024)Xue, Zheng, Hu, Liu, Chen, and Lou]{xue2024badragidentifyingvulnerabilitiesretrieval}
Xue, J., Zheng, M., Hu, Y., Liu, F., Chen, X., and Lou, Q.
\newblock Badrag: Identifying vulnerabilities in retrieval augmented generation of large language models, 2024.
\newblock URL \url{https://arxiv.org/abs/2406.00083}.

\bibitem[Yin et~al.(2018{\natexlab{a}})Yin, Chen, Kannan, and Bartlett]{trimmedmean}
Yin, D., Chen, Y., Kannan, R., and Bartlett, P.
\newblock {B}yzantine-robust distributed learning: Towards optimal statistical rates.
\newblock In Dy, J. and Krause, A. (eds.), \emph{Proceedings of the 35th International Conference on Machine Learning}, volume~80 of \emph{Proceedings of Machine Learning Research}, pp.\  5650--5659. PMLR, 10--15 Jul 2018{\natexlab{a}}.
\newblock URL \url{https://proceedings.mlr.press/v80/yin18a.html}.

\bibitem[Yin et~al.(2018{\natexlab{b}})Yin, Chen, Kannan, and Bartlett]{yin2018byzantine}
Yin, D., Chen, Y., Kannan, R., and Bartlett, P.
\newblock Byzantine-robust distributed learning: Towards optimal statistical rates.
\newblock In \emph{International Conference on Machine Learning}, pp.\  5650--5659. PMLR, 2018{\natexlab{b}}.

\bibitem[Zhou et~al.(2020)Zhou, Wu, Wang, and He]{biasvar}
Zhou, Y., Wu, J., Wang, H., and He, J.
\newblock Adversarial robustness through bias variance decomposition: A new perspective for federated learning.
\newblock \emph{arXiv preprint arXiv:2009.09026}, 2020.

\bibitem[Zizzo et~al.(2020)Zizzo, Rawat, Sinn, and Buesser]{FAT}
Zizzo, G., Rawat, A., Sinn, M., and Buesser, B.
\newblock Fat: Federated adversarial training.
\newblock \emph{arXiv preprint arXiv:2012.01791}, 2020.

\bibitem[Zizzo et~al.(2021)Zizzo, Rawat, Sinn, Maffeis, and Hankin]{certifiedfed}
Zizzo, G., Rawat, A., Sinn, M., Maffeis, S., and Hankin, C.
\newblock Certified federated adversarial training.
\newblock \emph{arXiv preprint arXiv:2112.10525}, 2021.

\end{thebibliography}
